\documentclass[11pt]{article}

\usepackage[final]{acl}

\usepackage{times}
\usepackage{latexsym}

\usepackage[T1]{fontenc}

\usepackage[utf8]{inputenc}

\usepackage{microtype}

\usepackage{inconsolata}

\usepackage{graphicx}
\usepackage{xcolor}
\usepackage{listings,booktabs,multirow,hyperref}
\usepackage[most]{tcolorbox}

\title{Beyond Calibration: Temporal LLM Confidence Assessment Across Languages and Local Settings}

\author{
 \textbf{Sahil Kale\textsuperscript{1}}
\\
 \textsuperscript{1}University of California, Los Angeles \\ CA, USA
\\
 \small{
   \textbf{Correspondence:} \href{mailto:sahilrkale@ucla.edu}{sahilrkale@ucla.edu}
 }
}

\begin{document}
\maketitle
\begin{abstract}

Large language models are expected to estimate their own uncertainty, with confidence signals used to guide retrieval, adaptive compute allocation, routing, and selective abstention. We investigate whether confidence signals measured at different temporal stages of problem solving encode distinct forms of epistemic information and remain reliable beyond English-centric settings. Inspired by classical meta-memory theory, we distinguish anticipatory pre-solve confidence (\emph{Feeling-of-Knowing}, FOK) from retrospective post-solve confidence (\emph{Judgement-of-Learning}, JOL). We evaluate these signals across two frontier LLMs, multiple reasoning domains, translated Marathi benchmarks, and a newly constructed benchmark of genuinely local Marathi factual questions. Across settings, the cheaper pre-solve signal remains weakly calibrated, while post-solve confidence generally provides stronger epistemic alignment despite its additional inference cost. Confidence behaviour transfers surprisingly well under direct language translation, with calibration trends largely preserved between English and Marathi, but degrades substantially on local Marathi knowledge, suggesting that linguistic transfer alone is insufficient for region-specific information needs. Finally, hallucinated responses maintain persistently elevated confidence after answer generation, highlighting important limitations of current LLM self-assessment mechanisms.

\end{abstract}

\section{Introduction}
Large language models (LLMs) are increasingly deployed across a wide range of real-world applications, including question answering, reasoning systems, retrieval-augmented generation (RAG), autonomous agents, and decision-support pipelines \cite{li-etal-2024-fundamental,cheng-etal-2025-realm}. Modern models contain substantial factual and procedural knowledge encoded within their parameters, while also leveraging external context through tools, retrieval systems, and long-context prompting. In many such systems, models are further expected to make decisions about their own uncertainty and knowledge limitations, a capability often framed as \textit{self-knowledge} \cite{kale-vrn-2025-line} or \textit{metacognition} \cite{ma-etal-2025-llm-metacognition}. These internal confidence estimates are increasingly used to influence downstream behaviours such as deciding whether to invoke tools, request additional user clarification, defer generation, allocate additional compute, or trigger retrieval mechanisms \cite{li-etal-2025-metacognition}. However, recent work has shown that LLM confidence signals are often poorly calibrated and only weakly aligned with actual correctness. This limits the efficiency, reliability, and cost-effectiveness of such confidence-driven decision policies \cite{kale2025knowrlteachinglanguagemodels}.

\begin{figure}[t]
    \centering
    \includegraphics[width=0.9\columnwidth]{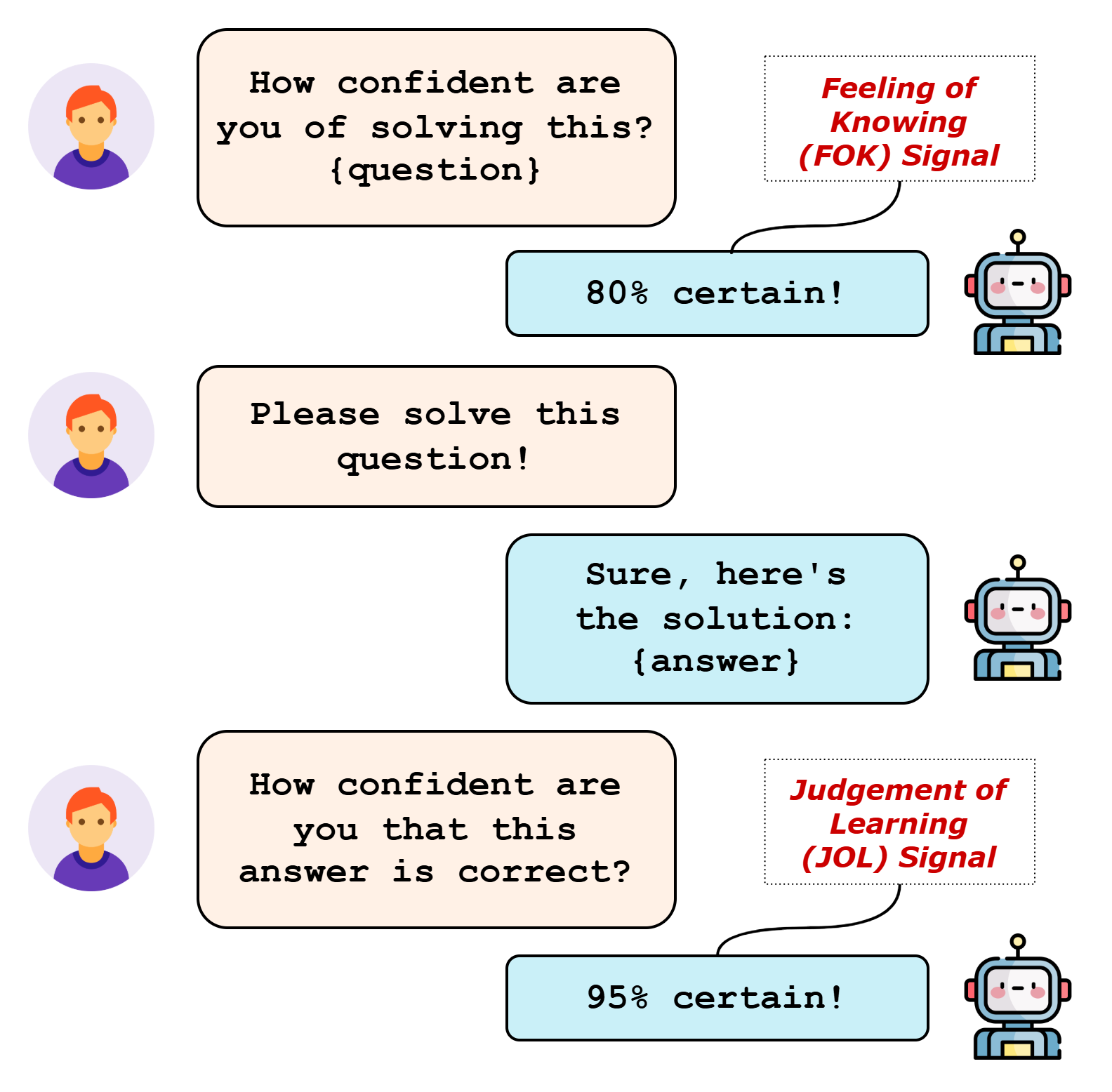}
    \caption{Illustration of pre-solve Feeling-of-Knowing (FOK) and post-solve Judgement-of-Learning (JOL) confidence signals in large language models.}
    \label{fig:fok_jol}
\end{figure}

While confidence calibration has been studied primarily in English-centric benchmarks, many practical deployments increasingly involve multilingual and underrepresented language settings where both model knowledge and evaluation resources are less abundant. Confidence estimates that appear reliable in high-resource settings may fail to transfer across languages or culturally specific knowledge domains, potentially leading to inappropriate retrieval decisions, overconfident hallucinations, or unsafe downstream behaviour. A particularly important open question is whether confidence calibration is primarily tied to language or to underlying knowledge availability. A model may retain similar confidence behaviour when questions are translated into another language, yet fail when confronted with genuinely local information that is sparsely represented in pre-training corpora and evaluation benchmarks. Distinguishing between these sources of distribution shift is crucial for deploying confidence-driven systems in multilingual settings.

In this work, we investigate whether confidence signals measured at different temporal stages of reasoning encode distinct forms of epistemic information, and also whether these carry across low-resource settings. Inspired by classical constructs from cognitive psychology and meta-memory research \cite{nelson1990metamemory,koriat1997monitoring}, we study two complementary signals: a pre-solve \emph{Feeling-of-Knowing} (FOK) signal, and a post-solve confidence signal similar to \emph{Judgement-of-Learning} (JOL). The FOK signal measures the model's estimated likelihood of solving a problem correctly before attempting to start reasoning towards a solution, capturing a cheap yet possibly useful anticipatory signal. In contrast, the JOL-style signal measures confidence in a provided reasoning path and solution after answer generation, reflecting retrospective evaluation expected to be costlier but better calibrated. This framing introduces a temporal distinction that is largely absent in existing LLM calibration work, where confidence is typically treated as a static scalar attached only to the final answer. In this paper, we instead ask whether epistemic self-assessment changes between pre-solve and post-solve states, and whether this shift itself carries useful information about model behaviour, reliability, and failure modes.

In essence, our work is guided by the following research questions: \textit{RQ1:} Is the pre-solve confidence signal (FOK) or the post-solve confidence signal (JOL) better calibrated to actual answer correctness, and can either signal support downstream decisions such as routing, retrieval triggering, or adaptive compute allocation? We also explore the combined effect of these signals in error types by answering another important question, \textit{RQ2:} Do strong values of FOK and JOL or delta trends correspond to distinct categories of model failure, such as knowledge gaps, reasoning failures, or hallucinated responses? In addition, we investigate a third research question motivated by multilingual deployment settings: \textit{RQ3:} Do temporal confidence signals transfer across low-resource languages, and how are they affected by genuinely local knowledge distributions that differ from standard English-centric benchmarks?

We operationalise FOK and JOL confidence signals using prompting-based self-assessment before and after reasoning generation across multiple LLM families and domains. We evaluate their calibration, cost-effectiveness, and transition dynamics using correctness-conditioned analyses and confidence calibration metrics. The main contributions of our work can be highlighted as follows: 

\begin{itemize}
    \item We introduce a novel temporal framing of LLM self-confidence by separating anticipatory pre-solve confidence from retrospective post-solve confidence and show that the cheaper pre-solve signal is generally insufficiently calibrated for practical deployment, while post-solve confidence is often more informative but not uniformly more reliable

    \item We provide one of the first analyses of temporal confidence calibration in two frontier LLMs, demonstrating that the practical utility of confidence signals depends strongly on reasoning domain and model family

    \item We extend temporal confidence evaluation to low-resource settings in Marathi, showing that confidence behaviour transfers relatively well across language translation but degrades substantially for genuinely local knowledge
\end{itemize}

\section{Related Work}

\subsection{Self-Knowledge and Confidence Estimation in LLMs}

Recent work has increasingly framed large language models as systems capable of exhibiting forms of self-knowledge, self-monitoring, or metacognitive behaviour \cite{Kale2025Mirage,podolak-verma-2025-read}. In this context, self-knowledge refers to a model's ability to estimate whether its own generated answer is likely to be correct or reliable \cite{li-etal-2025-metacognition}. 

Prior work has shown that LLMs can often distinguish between questions they are likely to answer correctly and those where they are uncertain, particularly under calibrated confidence or selective abstention prompting \cite{kale2025knowrlteachinglanguagemodels}. However, these signals remain imperfect, with models frequently exhibiting overconfidence, hallucination-prone certainty, and weak calibration under reasoning-heavy or distribution-shifted settings \cite{lin2022teaching,ren2026idontknowevaluating}. 

Despite these limitations, confidence-aware signals are increasingly important for applications such as retrieval triggering, adaptive compute allocation, selective generation, tool invocation, and human-in-the-loop systems \cite{Wang2026SKR,zhang2026stopfailoperationalcapability}. 

\subsection{Measuring Self-Knowledge and Confidence Signals}

Prior work has proposed a broad range of approaches for measuring confidence in language models. A major distinction concerns whether confidence is inferred from internal model representations (\textit{white-box} methods) or elicited through prompting and behavioural outputs (\textit{black-box} methods). White-box approaches typically rely on token probabilities, entropy, hidden-state representations, or calibration statistics derived directly from model distributions \cite{Kadavath2022,kissling2026selfawareknowledgeprobingevaluating}. In contrast, black-box methods estimate self-knowledge through verbalised confidence ratings, self-evaluation prompts, selective prediction, answer consistency, critique-and-revision behaviours, and confidence-guided abstention mechanisms \cite{xu-etal-2024-sayself,kale2025knowrlteachinglanguagemodels}.

Most prior approaches measure confidence only after answer generation, implicitly treating self-knowledge as a static retrospective signal attached to a completed response. In contrast, our work introduces a temporal perspective on confidence estimation by distinguishing between anticipatory pre-solve confidence and retrospective post-solve confidence.

\subsection{Metacognition, Distribution Shifts, and Underrepresented Languages}

A growing body of work has examined the robustness of LLM capabilities under distribution shifts arising from changes in task domains, reasoning requirements, languages, or cultural contexts \cite{moskvoretskii-etal-2025-adaptive,wiemerslage-wense-2025-improving}. While multilingual evaluation has received increasing attention across generation, reasoning, and factual knowledge tasks, comparatively little work has investigated whether models can accurately assess their own uncertainty in these settings \cite{durrani-etal-2025-editing}. This distinction is important because strong task performance does not necessarily imply well-calibrated confidence estimates. Our work contributes to analysis beyond English by evaluating temporal confidence signals in Marathi using both translated factual QA tasks and locally grounded knowledge questions designed to probe underrepresented linguistic and cultural contexts.

\section{Methodology}

Our methodology is designed to measure two temporally distinct confidence signals associated with problem solving in large language models: a pre-solve anticipatory confidence signal and a post-solve retrospective confidence signal. Broadly inspired by classical meta-memory constructs, we refer to these signals as \emph{Feeling-of-Knowing} (FOK) and \emph{Judgement-of-Learning} (JOL), respectively. Because our evaluation includes closed-source models accessible only through API interfaces, we use prompting-based verbal confidence estimates as the primary confidence signal, despite known limitations. The overall pipeline consists of three sequential stages shown in Figure \ref{fig:fok_jol}: (1) eliciting a pre-solve confidence estimate before reasoning begins, (2) generating an unconstrained answer to the task, and (3) eliciting a post-solve confidence estimate conditioned on the generated answer.

\subsection{Pre-Solve Confidence Signal (FOK)}
\begin{figure}[t]
    \centering
    \includegraphics[width=0.95\columnwidth]{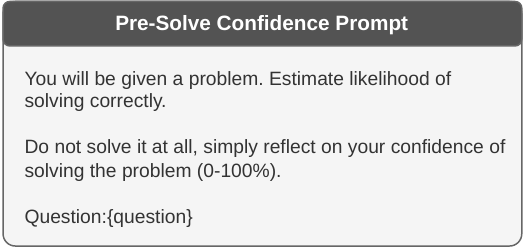}
    \caption{Prompt used for eliciting pre-solve confidence from LLMs}
    \label{fig:p1}
\end{figure}

To obtain the pre-solve confidence signal, the model is first shown only the input question and is asked to estimate the likelihood that it would solve the problem correctly without actually attempting the solution. The prompt explicitly instructs the model not to reason about or solve the task, and to instead provide only a simple confidence estimate. The prompt used for pre-solve confidence elicitation is shown in Figure~\ref{fig:p1}.

The scalar value returned by the model is treated as the FOK estimate. This signal captures the model's anticipatory belief regarding whether its internal knowledge and reasoning capabilities are sufficient to successfully solve the task prior to answer generation.

\subsection{Answer Generation}
After obtaining the pre-solve confidence estimate, the model is prompted again with the original question and asked to directly solve the task with chain-of-thought reasoning. No additional confidence instructions, calibration prompts, or answer formatting requirements are imposed during this stage. This design choice ensures that the generated reasoning trajectory more closely reflects the model's natural inference behaviour under standard prompting conditions.

The generated answer is then stored and used as contextual input for retrospective confidence estimation in the next stage.

\subsection{Post-Solve Confidence Signal (JOL)}

\begin{figure}[t]
    \centering
    \includegraphics[width=0.95\columnwidth]{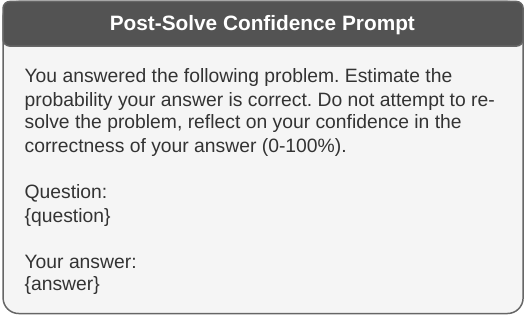}
    \caption{Prompt used for eliciting post-solve confidence from LLMs}
    \label{fig:p2}
\end{figure}

To obtain the post-solve confidence signal, the model is shown both the original question and its previously generated answer. It is then asked to estimate the probability that its answer is correct. Unlike the pre-solve stage, this confidence estimate is conditioned on the model's completed reasoning trajectory and final output. The post-solve confidence prompt is shown in Figure \ref{fig:p2}.

The scalar value returned by the model is then used as a JOL-style retrospective confidence estimate. This signal reflects self-evaluation conditioned on the produced answer and reasoning path rather than on the question alone.

Following these steps, for each problem instance, we therefore obtain three core variables: a pre-solve confidence estimate (FOK), a generated answer, and a post-solve confidence estimate (JOL). The generated answer is independently evaluated against reference solutions to determine correctness labels. We then analyse confidence calibration estimates and cost-effectiveness of these signals ahead. 

For multilingual experiments, the same three-stage procedure is retained without modification. Confidence elicitation, answer generation, and post-solve self-assessment prompts are translated into Marathi while preserving their original structure and intent. This design allows us to study whether temporal confidence signals remain stable under linguistic transfer without changing the underlying confidence estimation framework.

\section{Experimental Setup}

\subsection{Models}

We evaluate our framework using two frontier large language models: \texttt{GPT-5.4 (gpt-5.4-2026-03-05)} and \texttt{Claude Sonnet 4.6}. Models were accessed through the official OpenAI and Anthropic APIs. All experiments use deterministic decoding with temperature set to $0$, while other generation parameters remain at default settings. Full details are provided in Section \ref{app:models} in the Appendix.

\subsection{Datasets}

We evaluate across four domains probing mathematical reasoning, logical reasoning, factual recall, and low-resource multilingual factual recall.

\subsubsection{Mathematical Reasoning}

For mathematical reasoning, we use 500 randomly sampled questions from the Mathematics Stack Exchange split of \texttt{RealMath} \cite{realmath}. The dataset contains non-trivial user-authored mathematical problems with verified solutions and diverse reasoning structures, making it suitable for studying confidence calibration under open-ended mathematical reasoning.

\subsubsection{Logical Reasoning}

For logical reasoning, we use 400 samples from \texttt{RiddleBench} \cite{riddlebench}, a benchmark of challenging natural-language puzzles involving deduction, constraint satisfaction, spatial reasoning, and multi-step inference. These tasks provide a cognitively demanding setting for evaluating confidence behaviour under complex reasoning.

\subsubsection{Factual Recall}

For factual recall, we use 750 randomly sampled questions from \texttt{TriviaQA} \cite{JoshiTriviaQA2017}, a large-scale benchmark of naturally authored trivia questions with verified answers. Unlike the reasoning datasets above, TriviaQA primarily evaluates factual retrieval and parametric knowledge recall.

\subsubsection{Low-Resource Settings}

To evaluate temporal confidence signals in an underrepresented language setting, we construct two Marathi factual-recall splits.

The first split (\textit{Marathi-Translated}) is derived directly by translating TriviaQA, RealMath and RiddleBench to a low-resource language, Marathi with each split retaining the same number of examples as its English counterpart (750 TriviaQA, 500 RealMath, 400 RiddleBench). Questions and reference answers are translated from English to Marathi using the Google Translate API while preserving the original task structure and evaluation protocol. Two graduate-level native Marathi speakers reviewed translated prompts and dataset samples, performed spot-checks throughout the corpus, and verified semantic equivalence with the original English questions. This split isolates language transfer while keeping knowledge requirements unchanged. Additional details are in Section \ref{app:marathi-glo} in the Appendix.

The second split (\textit{Marathi-Local}) evaluates genuinely local knowledge. Unlike translated benchmarks, these questions require region-specific information that is less frequently represented in English-language resources and often available primarily through Marathi-language sources. Three native Marathi speakers manually authored 201 factual questions based on local events and public information reported before model knowledge cut-off dates using trusted Marathi news outlets such as Zee News Marathi, Maharashtra Times, and ABP Majha. Each question was paired with supporting source URLs for verification, while reference answers and correctness assessments were independently reviewed by two separate native speakers with agreement of Cohen's $\kappa=0.82$. Additional details and examples are in Section \ref{app:marathi-local} in the Appendix.

\subsection{Evaluation Protocol}

Answers are evaluated using benchmark-specific matching procedures to obtain correctness labels. We report Expected Calibration Error (ECE) \cite{ece-brier}, Brier Score \cite{ece-brier}, and AUROC. ECE measures the gap between predicted confidence and empirical correctness, Brier Score evaluates the quality of probabilistic confidence estimates, and AUROC measures how effectively confidence scores distinguish correct from incorrect predictions. We note that AUROC evaluates discrimination independently of calibration and can remain high even when ECE or Brier Score indicate poor calibration, but we report all three because they capture complementary properties of confidence quality. Reliability curves are additionally used to visualise calibration behaviour. Uncertainty estimates are computed using 2000 bootstrap resamples and reported as 95\% confidence intervals. We further report cost-benefit analyses and confidence trends across error categories.

For the Marathi-Translated split, evaluation was adapted to support valid multilingual answer variations. Each example contains both the original English reference answer and its Marathi translation. Predictions are considered correct if they match either reference set under the benchmark evaluation procedure, reducing false negatives caused by transliteration differences, bilingual generations, or alternative entity forms. Native-speaker spot checks were conducted to verify evaluation quality.

For the Marathi-Local split, no benchmark labels are available. Answer correctness is therefore assessed manually by native Marathi speakers using the associated source material. Following prior factuality evaluation work, correctness is scored on a continuous scale between 0 and 1, where 1 denotes a fully correct answer, 0 an incorrect answer, and intermediate values partially correct responses (AUROC computed with threshold of 0.5). These scores are subsequently used when computing calibration and confidence metrics.

\begin{figure}[t]
    \centering
    \includegraphics[width=0.9\columnwidth]{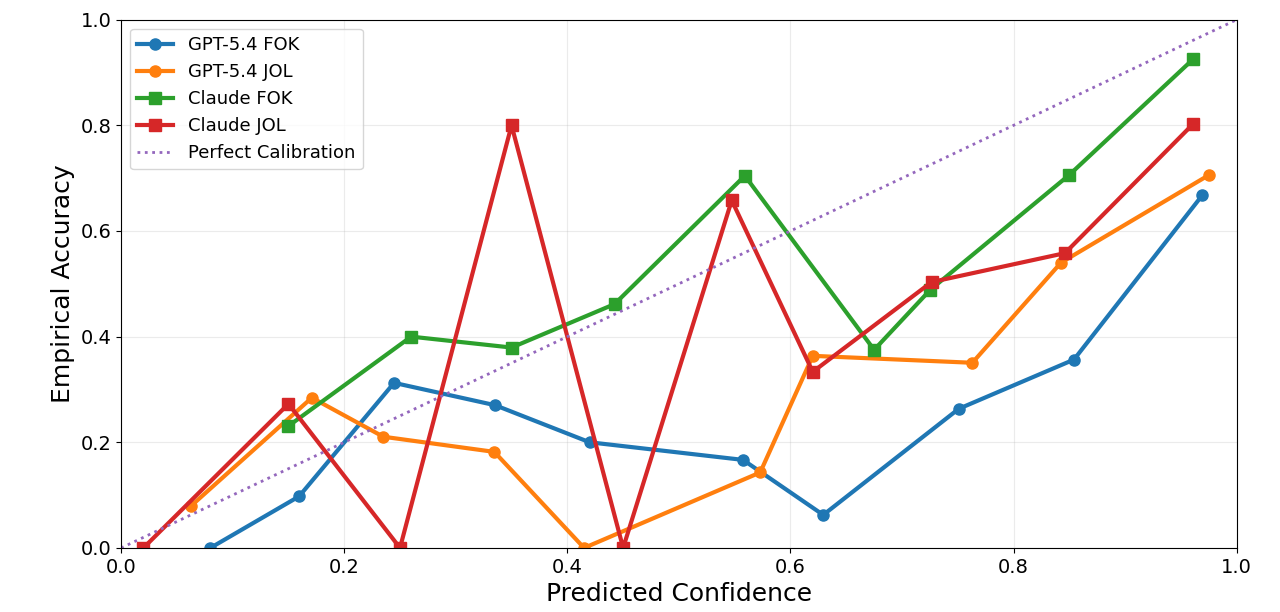}
    \caption{Aggregate reliability curves comparing pre-solve (FOK) and post-solve (JOL) confidence calibration across models and domains}
    \label{fig:curves}
\end{figure}

\begin{table*}[t]
\centering

\scriptsize
\renewcommand{\arraystretch}{1.05}

\resizebox{\textwidth}{!}{
\begin{tabular}{llcccccc}
\toprule
\multirow{2}{*}{\textbf{Model}} &
\multirow{2}{*}{\textbf{Domain}} &
\multicolumn{3}{c}{\textbf{Pre-Solve (FOK)}} &
\multicolumn{3}{c}{\textbf{Post-Solve (JOL)}} \\

\cmidrule(lr){3-5}
\cmidrule(lr){6-8}

&
& \textbf{ECE}$\downarrow$
& \textbf{Brier}$\downarrow$
& \textbf{AUROC}$\uparrow$
& \textbf{ECE}$\downarrow$
& \textbf{Brier}$\downarrow$
& \textbf{AUROC}$\uparrow$ \\

\midrule

\multirow{3}{*}{GPT-5.4}
& TriviaQA & $0.195\pm0.027$ & $0.209\pm0.028$ & $0.567\pm0.053$ & $0.194\pm0.021$ & $0.203\pm0.025$ & $0.606\pm0.051$ \\
& Math     & $0.381\pm0.041$ & $0.366\pm0.037$ & $0.719\pm0.047$ & $0.315\pm0.045$ & $0.330\pm0.037$ & $0.751\pm0.048$ \\
& Logic    & $0.483\pm0.039$ & $0.469\pm0.035$ & $0.679\pm0.055$ & $0.286\pm0.041$ & $0.313\pm0.036$ & $0.690\pm0.049$ \\

\midrule

\multirow{3}{*}{Claude Sonnet 4.6}
& TriviaQA & $0.071\pm0.023$ & $0.119\pm0.019$ & $0.656\pm0.055$ & $0.077\pm0.024$ & $0.114\pm0.020$ & $0.703\pm0.046$ \\
& Math     & $0.188\pm0.061$ & $0.231\pm0.051$ & $0.500\pm0.092$ & $0.108\pm0.064$ & $0.140\pm0.047$ & $0.791\pm0.085$ \\
& Logic    & $0.341\pm0.051$ & $0.316\pm0.029$ & $0.767\pm0.057$ & $0.472\pm0.053$ & $0.456\pm0.041$ & $0.695\pm0.049$ \\

\bottomrule
\end{tabular}
}
\caption{Calibration and discrimination performance (95\% confidence intervals) across domains and temporal stages}
\label{tab:main_results}
\end{table*}

\section{Results}

Table~\ref{tab:main_results} reports calibration and discrimination performance for pre-solve (FOK) and post-solve (JOL) confidence signals across models and domains. Reported values are means with 95\% bootstrap confidence intervals. Figure~\ref{fig:curves} shows aggregate reliability curves. We evaluate calibration using Expected Calibration Error (ECE) and Brier Score, where lower values indicate stronger alignment between confidence and correctness, and discrimination using AUROC.

\subsection{Calibration Analysis}

\subsubsection{\textbf{Post-solve confidence generally provides stronger epistemic alignment}}

Across most domains, JOL-style confidence estimates exhibit lower calibration error and stronger discrimination than FOK estimates, although gains are not uniform and remain modest for factual recall. This trend is particularly pronounced in mathematics, where Claude Sonnet 4.6 improves from near-random discrimination ($0.500 \pm 0.092$) to strong discrimination ($0.791 \pm 0.085$). The consistently weaker calibration of the cheaper pre-solve signal suggests that FOK alone is insufficiently reliable for downstream decisions such as retrieval triggering, model routing, or adaptive compute allocation. Instead, models appear to require answer generation before producing confidence estimates that better reflect correctness.

\subsubsection{\textbf{Reasoning domains exhibit substantially higher calibration instability than factual recall}}

Both models remain notably miscalibrated on mathematics and logical reasoning despite moderate discrimination performance. Logic reasoning in particular exhibits consistently high ECE values, with GPT-5.4 reaching a pre-solve ECE of $0.483 \pm 0.039$ and Claude Sonnet 4.6 a post-solve ECE of $0.472 \pm 0.053$. In contrast, TriviaQA shows lower calibration error and smaller temporal shifts between FOK and JOL. This suggests that retrospective confidence adjustment is more useful for multi-step reasoning tasks than for direct factual retrieval, where confidence remains relatively stable before and after answer generation.

\subsubsection{\textbf{Temporal confidence transitions reveal domain-dependent failure behaviour}}

The transition from FOK to JOL is not uniformly beneficial. While post-solve confidence often improves calibration and ranking quality, certain domains exhibit confidence inflation after reasoning. Most notably, Claude Sonnet 4.6 on logic reasoning degrades from an ECE of $0.341$ to $0.472$, while AUROC falls from $0.767$ to $0.695$. The corresponding reliability curves are also less stable after reasoning. These results suggest that generating extended reasoning traces can sometimes induce false certainty rather than improved self-knowledge.

\begin{figure}[t]
    \centering
    \includegraphics[width=0.9\columnwidth]{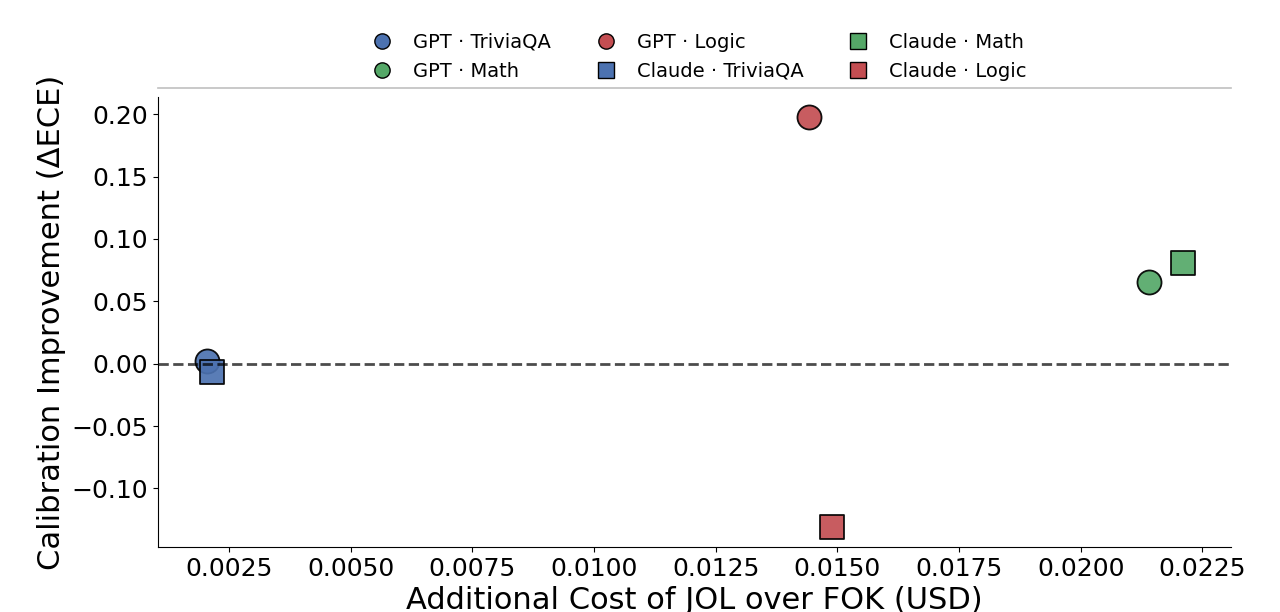}
    \caption{Calibration improvement versus additional inference cost for post-solve confidence (JOL) relative to pre-solve confidence (FOK) across models and domains.}
    \label{fig:cost}
\end{figure}

\subsection{Cost-Benefit Analysis}

We next examine the trade-off between the additional inference cost required to obtain JOL confidence and the calibration gains achieved. FOK cost includes only the confidence estimation prompt, whereas JOL additionally requires answer generation. We note this cost comparison assumes an application where answer generation is not otherwise required (e.g., pre-generation retrieval routing). In settings where an answer must be produced regardless (e.g., post-hoc abstention or answer verification), the marginal cost of JOL is limited to the additional confidence-elicitation query. Costs were estimated using average dataset lengths and provider pricing rates as of May 2026. Positive $\Delta$ECE values indicate improved calibration from JOL relative to FOK, as shown in Figure~\ref{fig:cost}.

\subsubsection{\textbf{JOL is primarily cost-effective in structured reasoning domains}}

For GPT-5.4, JOL consistently improves calibration, with the largest gains occurring in Logic ($\Delta$ECE = 0.198). Mathematics also shows meaningful improvements for both models, whereas TriviaQA exhibits only marginal gains. These results suggest that post-solve confidence becomes most useful as reasoning complexity increases and can justify additional inference cost in structured problem-solving settings.

\subsubsection{\textbf{Additional reasoning does not always improve self-assessment}}

TriviaQA shows minimal calibration improvement from JOL and slight degradation for Claude Sonnet 4.6, suggesting that the cheaper FOK signal may already be sufficient for factual recall. More notably, Claude on Logic becomes substantially less calibrated despite higher inference cost, indicating that additional reasoning can sometimes amplify miscalibration rather than improve confidence reliability.

\begin{figure}[t]
    \centering
    \includegraphics[width=0.9\columnwidth]{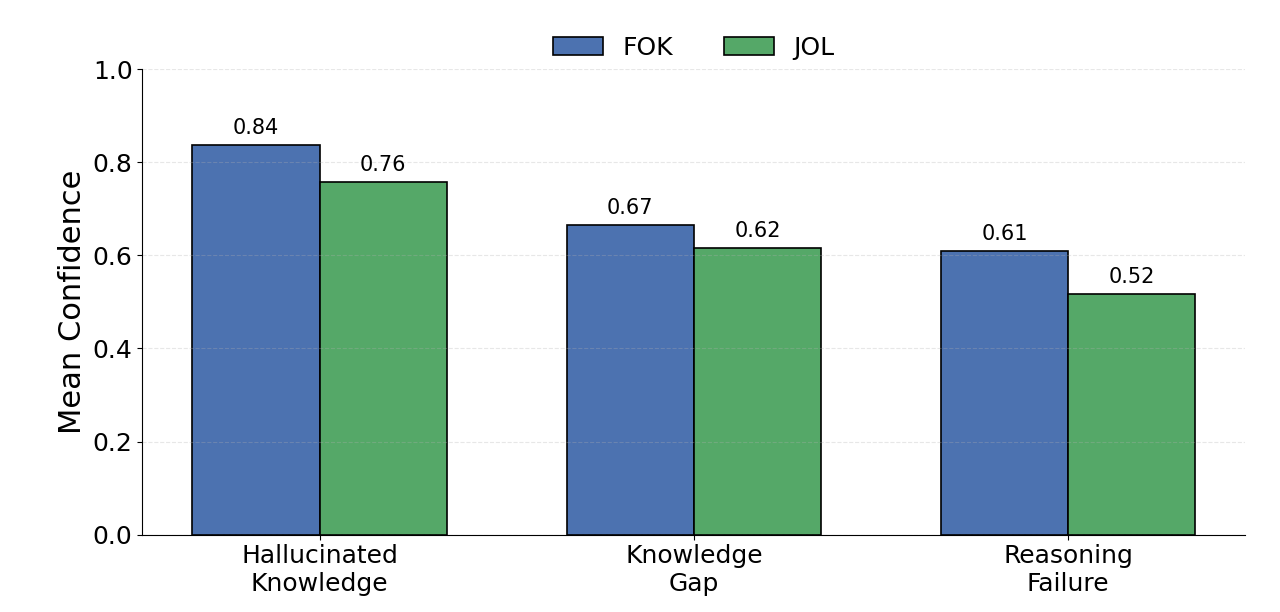}
    \caption{Mean pre-solve (FOK) and post-solve (JOL) confidence across annotated error categories.}
    \label{fig:error}
\end{figure}

\subsection{Error Type Analysis}

To investigate whether confidence patterns correspond to distinct failure mechanisms, we manually analysed 200 incorrect responses sampled across models and domains. Two graduate-level annotators independently reviewed all samples, achieving a Cohen's $\kappa$ of 0.92. Disagreements were subsequently resolved through discussion. Errors were assigned to one of three categories:

\begin{itemize}
    \item \textit{Knowledge gap:} The model lacked the required information or explicitly hedged between alternatives.
    \item \textit{Hallucinated knowledge:} The model confidently generated fabricated facts, entities, or incorrect theorem usage.
    \item \textit{Reasoning failure:} The answer was incorrect due to flawed reasoning despite the absence of fabricated knowledge.
\end{itemize}

Absolute confidence levels proved more informative than confidence transitions for distinguishing failure categories. Confidence deltas exhibited high variance and did not significantly separate error types ($p = 0.366$, Kruskal-Wallis test). Nevertheless, clear differences emerge in Figure~\ref{fig:error}, where confidence generally decreases from FOK to JOL, consistent with the calibration improvements observed earlier.

\subsubsection{\textbf{Hallucinations exhibit persistently elevated confidence}}

Hallucinated responses showed the highest confidence both before and after reasoning (FOK: 0.837, JOL: 0.758) despite being incorrect. This suggests that hallucinations are associated less with confidence increases during reasoning and more with overconfident internal representations that remain stable throughout answer generation, consistent with prior hallucination research \cite{simhi-etal-2025-trust,hallu-calib}.

\subsubsection{\textbf{Reasoning failures show lower and more unstable confidence patterns}}

Reasoning failures exhibited the lowest confidence levels alongside substantial variance in confidence transitions, indicating that models partially recognise reasoning difficulty even when failing. Knowledge gaps occupied an intermediate regime, where models often began uncertain but also decreased confidence after producing an answer.

\subsection{Multilingual Transfer and Local Knowledge}

\subsubsection{\textbf{Temporal confidence signals largely transfer across language translation}}

Direct translation of benchmark questions into Marathi (Marathi-Translated split) produces only modest changes in calibration and discrimination performance relative to the original English benchmarks, as seen in Table~\ref{tab:translation_transfer}. Across models and domains, most ECE differences remain small and within bootstrap uncertainty intervals. GPT-5.4 exhibits only minor calibration changes after translation, and temporal confidence dynamics remain largely unchanged across mathematical, logical, and factual tasks. Claude Sonnet 4.6 shows somewhat larger degradation for factual recall, but the overall pattern of stronger post-solve confidence discrimination persists.

These findings suggest that the temporal confidence mechanisms captured by FOK and JOL are relatively robust to language transfer. Translating tasks into an underrepresented language does not substantially alter how these frontier models assess their own uncertainty.

\subsubsection{\textbf{Locally grounded knowledge produces substantially larger calibration failures}}

A markedly different picture emerges on the Marathi-Local split. Both models exhibit substantially higher calibration error and weaker discrimination than on either English or translated Marathi benchmarks (Table~\ref{tab:local}).

For GPT-5.4, pre-solve ECE increases from 0.195 on English TriviaQA and 0.205 on translated Marathi TriviaQA to 0.421 on Marathi-Local questions. Claude Sonnet 4.6 exhibits a similar increase from 0.071 and 0.122 to 0.378. AUROC values simultaneously approach near-random ranking behaviour, falling to 0.521 and 0.558 respectively.

These results indicate that calibration degrades far more strongly under local knowledge shifts than under language shifts alone. While models largely preserve self-assessment behaviour across languages, they remain substantially less reliable when confronted with region-specific information that is sparsely represented in training data and retrieval resources.

\begin{table}[t]
\centering

\small
\begin{tabular}{llcc}
\toprule
Model & Domain & FOK $\Delta$ECE & JOL $\Delta$ECE \\
\midrule

GPT-5.4
& TriviaQA & +0.010 & -0.005 \\
& Math     & +0.030 & -0.016 \\
& Logic    & +0.028 & +0.019 \\

\midrule

Claude
& TriviaQA & +0.051 & +0.045 \\
& Math     & +0.004 & +0.016 \\
& Logic    & -0.009 & +0.040 \\

\bottomrule
\end{tabular}
\caption{Change in calibration error ($\Delta$ECE) after translation from English to Marathi. Positive values indicate worse calibration in Marathi.}
\label{tab:translation_transfer}
\end{table}

\begin{table}[t]
\centering

\small
\begin{tabular}{lccc}
\toprule
Model & Signal & ECE & AUROC \\
\midrule
GPT-5.4 & FOK & 0.421 & 0.521 \\
GPT-5.4 & JOL & 0.287 & 0.579 \\
Claude & FOK & 0.378 & 0.558 \\
Claude & JOL & 0.255 & 0.616 \\
\bottomrule
\end{tabular}
\caption{Confidence calibration on the Marathi-Local benchmark.}
\label{tab:local}
\end{table}

\section{Conclusion and Future Work}

In this work, we introduced a temporal framing of confidence estimation in large language models by separating pre-solve (FOK) and post-solve (JOL) confidence signals. Across models and reasoning domains, post-solve confidence generally provides stronger calibration and discrimination than the cheaper pre-solve signal, although gains depend strongly on the model and task. In Marathi, confidence behaviour transfers well under direct language translation, but calibration degrades substantially on genuinely local knowledge questions. Hallucinated responses also maintain persistently elevated confidence, highlighting limitations of current self-assessment mechanisms. Overall, confidence-aware systems should account for both temporal stage and knowledge domain when using confidence for retrieval, routing, or adaptive compute decisions.

Future work can extend this framework to additional low-resource languages, multimodal settings, agentic systems, and model families beyond the two evaluated here. Because Marathi-Local and Marathi-Translated differ along several axes besides locality, the calibration gap cannot be attributed to locality alone; a matched benchmark controlling for these factors would help isolate its effect. It would also be valuable to examine whether white-box uncertainty signals, such as token probabilities and representation-based confidence, exhibit similar temporal dynamics. More broadly, our findings motivate studying calibration not only across languages but also across local knowledge distributions, where linguistic transfer alone may be insufficient for reliable self-assessment.

\bibliography{custom}

@inproceedings{cheng-etal-2025-realm,
    title = "{REALM}: A Dataset of Real-World {LLM} Use Cases",
    author = "Cheng, Jingwen  and
      Ghate, Kshitish  and
      Hua, Wenyue  and
      Wang, William Yang  and
      Shen, Hong  and
      Fang, Fei",
    editor = "Che, Wanxiang  and
      Nabende, Joyce  and
      Shutova, Ekaterina  and
      Pilehvar, Mohammad Taher",
    booktitle = "Findings of the Association for Computational Linguistics: ACL 2025",
    month = jul,
    year = "2025",
    address = "Vienna, Austria",
    publisher = "Association for Computational Linguistics",
    url = "https://aclanthology.org/2025.findings-acl.437/",
    doi = "10.18653/v1/2025.findings-acl.437",
    pages = "8331--8341",
    ISBN = "979-8-89176-256-5"
}

@inproceedings{li-etal-2024-fundamental,
    title = "Fundamental Capabilities of Large Language Models and their Applications in Domain Scenarios: A Survey",
    author = "Li, Jiawei  and
      Yang, Yizhe  and
      Bai, Yu  and
      Zhou, Xiaofeng  and
      Li, Yinghao  and
      Sun, Huashan  and
      Liu, Yuhang  and
      Si, Xingpeng  and
      Ye, Yuhao  and
      Wu, Yixiao  and
      Lin, Yiguan  and
      Xu, Bin  and
      Ren, Bowen  and
      Feng, Chong  and
      Gao, Yang  and
      Huang, Heyan",
    editor = "Ku, Lun-Wei  and
      Martins, Andre  and
      Srikumar, Vivek",
    booktitle = "Proceedings of the 62nd Annual Meeting of the Association for Computational Linguistics (Volume 1: Long Papers)",
    month = aug,
    year = "2024",
    address = "Bangkok, Thailand",
    publisher = "Association for Computational Linguistics",
    url = "https://aclanthology.org/2024.acl-long.599/",
    doi = "10.18653/v1/2024.acl-long.599",
    pages = "11116--11141"
}

@inproceedings{kale-vrn-2025-line,
    title = "Line of Duty: Evaluating {LLM} Self-Knowledge via Consistency in Feasibility Boundaries",
    author = "Kale, Sahil  and
      Nadadur, Vijaykant",
    editor = "Cao, Trista  and
      Das, Anubrata  and
      Kumarage, Tharindu  and
      Wan, Yixin  and
      Krishna, Satyapriya  and
      Mehrabi, Ninareh  and
      Dhamala, Jwala  and
      Ramakrishna, Anil  and
      Galystan, Aram  and
      Kumar, Anoop  and
      Gupta, Rahul  and
      Chang, Kai-Wei",
    booktitle = "Proceedings of the 5th Workshop on Trustworthy NLP (TrustNLP 2025)",
    month = may,
    year = "2025",
    address = "Albuquerque, New Mexico",
    publisher = "Association for Computational Linguistics",
    url = "https://aclanthology.org/2025.trustnlp-main.10/",
    doi = "10.18653/v1/2025.trustnlp-main.10",
    pages = "127--140",
    ISBN = "979-8-89176-233-6"
}

@inproceedings{li-etal-2025-metacognition,
    title = "Adaptive Tool Use in Large Language Models with Meta-Cognition Trigger",
    author = "Li, Wenjun  and
      Li, Dexun  and
      Dong, Kuicai  and
      Zhang, Cong  and
      Zhang, Hao  and
      Liu, Weiwen  and
      Wang, Yasheng  and
      Tang, Ruiming  and
      Liu, Yong",
    editor = "Che, Wanxiang  and
      Nabende, Joyce  and
      Shutova, Ekaterina  and
      Pilehvar, Mohammad Taher",
    booktitle = "Proceedings of the 63rd Annual Meeting of the Association for Computational Linguistics (Volume 1: Long Papers)",
    month = jul,
    year = "2025",
    address = "Vienna, Austria",
    publisher = "Association for Computational Linguistics",
    url = "https://aclanthology.org/2025.acl-long.655/",
    doi = "10.18653/v1/2025.acl-long.655",
    pages = "13346--13370",
    ISBN = "979-8-89176-251-0"
}

@inproceedings{ma-etal-2025-llm-metacognition,
    title = "Large Language Models Have Intrinsic Meta-Cognition, but Need a Good Lens",
    author = "Ma, Ziyang  and
      Yuan, Qingyue  and
      Wang, Zhenglin  and
      Zhou, Deyu",
    editor = "Christodoulopoulos, Christos  and
      Chakraborty, Tanmoy  and
      Rose, Carolyn  and
      Peng, Violet",
    booktitle = "Proceedings of the 2025 Conference on Empirical Methods in Natural Language Processing",
    month = nov,
    year = "2025",
    address = "Suzhou, China",
    publisher = "Association for Computational Linguistics",
    url = "https://aclanthology.org/2025.emnlp-main.171/",
    doi = "10.18653/v1/2025.emnlp-main.171",
    pages = "3460--3477",
    ISBN = "979-8-89176-332-6"
}

@misc{kale2025knowrlteachinglanguagemodels,
      title={KnowRL: Teaching Language Models to Know What They Know}, 
      author={Sahil Kale and Devendra Singh Dhami},
      year={2025},
      eprint={2510.11407},
      archivePrefix={arXiv},
      primaryClass={cs.CL},
      url={https://arxiv.org/abs/2510.11407}, 
}

@misc{lin2022teaching,
      title={Teaching Models to Express Their Uncertainty in Words}, 
      author={Stephanie Lin and Jacob Hilton and Owain Evans},
      year={2022},
      eprint={2205.14334},
      archivePrefix={arXiv},
      primaryClass={cs.CL},
      url={https://arxiv.org/abs/2205.14334}, 
}

@incollection{nelson1990metamemory,
title = {Metamemory: A Theoretical Framework and New Findings},
editor = {Gordon H. Bower},
series = {Psychology of Learning and Motivation},
publisher = {Academic Press},
volume = {26},
pages = {125-173},
year = {1990},
issn = {0079-7421},
doi = {https://doi.org/10.1016/S0079-7421(08)60053-5},
url = {https://www.sciencedirect.com/science/article/pii/S0079742108600535},
author = {Thomas O. Nelson}
}

@article{koriat1997monitoring,
  author    = {Asher Koriat},
  title     = {Monitoring One's Own Knowledge During Study: A Cue-Utilization Approach to Judgments of Learning},
  journal   = {Journal of Experimental Psychology: General},
  year      = {1997},
  volume    = {126},
  number    = {4},
  pages     = {349--370},
  doi       = {10.1037/0096-3445.126.4.349},
  publisher = {American Psychological Association}
}

@inproceedings{podolak-verma-2025-read,
    title = "Read Your Own Mind: Reasoning Helps Surface Self-Confidence Signals in {LLM}s",
    author = "Podolak, Jakub  and
      Verma, Rajeev",
    editor = {Eikema, Bryan  and
      V{\'a}zquez, Ra{\'u}l  and
      Berant, Jonathan  and
      de Marneffe, Marie-Catherine  and
      Plank, Barbara  and
      Shelmanov, Artem  and
      Swayamdipta, Swabha  and
      Tiedemann, J{\"o}rg  and
      Zerva, Chrysoula  and
      Aziz, Wilker},
    booktitle = "Proceedings of the 2nd Workshop on Uncertainty-Aware NLP (UncertaiNLP 2025)",
    month = nov,
    year = "2025",
    address = "Suzhou, China",
    publisher = "Association for Computational Linguistics",
    url = "https://aclanthology.org/2025.uncertainlp-main.21/",
    doi = "10.18653/v1/2025.uncertainlp-main.21",
    pages = "247--258",
    ISBN = "979-8-89176-349-4"
}

@article{Kale2025Mirage,
  author    = {Sahil Kale},
  title     = {Mirage of Mastery: Memorization Tricks LLMs into Artificially Inflated Self-Knowledge},
  journal   = {arXiv preprint arXiv:2506.18998},
  year      = {2025},
  doi       = {10.48550/arXiv.2506.18998},
  url       = {https://arxiv.org/abs/2506.18998}
}

@misc{ren2026idontknowevaluating,
      title={Beyond "I Don't Know": Evaluating LLM Self-Awareness in Discriminating Data and Model Uncertainty}, 
      author={Jingyi Ren and Ante Wang and Yunghwei Lai and Xiaolong Wang and Linlu Gong and Weitao Li and Weizhi Ma and Yang Liu},
      year={2026},
      eprint={2604.17293},
      archivePrefix={arXiv},
      primaryClass={cs.CL},
      url={https://arxiv.org/abs/2604.17293}, 
}

@inproceedings{ece-brier,
author = {Naeini, Mahdi Pakdaman and Cooper, Gregory F. and Hauskrecht, Milos},
title = {Obtaining well calibrated probabilities using bayesian binning},
year = {2015},
isbn = {0262511290},
publisher = {AAAI Press},
booktitle = {Proceedings of the Twenty-Ninth AAAI Conference on Artificial Intelligence},
pages = {2901–2907},
numpages = {7},
location = {Austin, Texas},
series = {AAAI'15}
}

@misc{zhang2026stopfailoperationalcapability,
      title={Stop Before You Fail: Operational Capability Boundaries for Mitigating Unproductive Reasoning in Large Reasoning Models}, 
      author={Qingjie Zhang and Yujia Fu and Yang Wang and Liu Yan and Tao Wei and Ke Xu and Minlie Huang and Han Qiu},
      year={2026},
      eprint={2509.24711},
      archivePrefix={arXiv},
      primaryClass={cs.AI},
      url={https://arxiv.org/abs/2509.24711}, 
}

@misc{kissling2026selfawareknowledgeprobingevaluating,
      title={Self-Aware Knowledge Probing: Evaluating Language Models' Relational Knowledge through Confidence Calibration}, 
      author={Christopher Kissling and Elena Merdjanovska and Alan Akbik},
      year={2026},
      eprint={2601.18901},
      archivePrefix={arXiv},
      primaryClass={cs.CL},
      url={https://arxiv.org/abs/2601.18901}, 
}

@inproceedings{xu-etal-2024-sayself,
    title = "{S}ay{S}elf: Teaching {LLM}s to Express Confidence with Self-Reflective Rationales",
    author = "Xu, Tianyang  and
      Wu, Shujin  and
      Diao, Shizhe  and
      Liu, Xiaoze  and
      Wang, Xingyao  and
      Chen, Yangyi  and
      Gao, Jing",
    editor = "Al-Onaizan, Yaser  and
      Bansal, Mohit  and
      Chen, Yun-Nung",
    booktitle = "Proceedings of the 2024 Conference on Empirical Methods in Natural Language Processing",
    month = nov,
    year = "2024",
    address = "Miami, Florida, USA",
    publisher = "Association for Computational Linguistics",
    url = "https://aclanthology.org/2024.emnlp-main.343/",
    doi = "10.18653/v1/2024.emnlp-main.343",
    pages = "5985--5998"
}

@misc{realmath,
      title={RealMath: A Continuous Benchmark for Evaluating Language Models on Research-Level Mathematics}, 
      author={Jie Zhang and Cezara Petrui and Kristina Nikolić and Florian Tramèr},
      year={2025},
      eprint={2505.12575},
      archivePrefix={arXiv},
      primaryClass={cs.AI},
      url={https://arxiv.org/abs/2505.12575}, 
}

@misc{riddlebench,
      title={RiddleBench: A New Generative Reasoning Benchmark for LLMs}, 
      author={Deepon Halder and Alan Saji and Thanmay Jayakumar and Ratish Puduppully and Anoop Kunchukuttan and Raj Dabre},
      year={2025},
      eprint={2510.24932},
      archivePrefix={arXiv},
      primaryClass={cs.CL},
      url={https://arxiv.org/abs/2510.24932}, 
}

@InProceedings{JoshiTriviaQA2017,
     author = {Joshi, Mandar and Choi, Eunsol and Weld, Daniel S. and Zettlemoyer, Luke},
     title = {TriviaQA: A Large Scale Distantly Supervised Challenge Dataset for Reading Comprehension},
     booktitle = {Proceedings of the 55th Annual Meeting of the Association for Computational Linguistics},
     month = {July},
     year = {2017},
     address = {Vancouver, Canada},
     publisher = {Association for Computational Linguistics},
}

@inproceedings{simhi-etal-2025-trust,
    title = "Trust Me, {I}{'}m Wrong: {LLM}s Hallucinate with Certainty Despite Knowing the Answer",
    author = "Simhi, Adi  and
      Itzhak, Itay  and
      Barez, Fazl  and
      Stanovsky, Gabriel  and
      Belinkov, Yonatan",
    editor = "Christodoulopoulos, Christos  and
      Chakraborty, Tanmoy  and
      Rose, Carolyn  and
      Peng, Violet",
    booktitle = "Findings of the Association for Computational Linguistics: EMNLP 2025",
    month = nov,
    year = "2025",
    address = "Suzhou, China",
    publisher = "Association for Computational Linguistics",
    url = "https://aclanthology.org/2025.findings-emnlp.792/",
    doi = "10.18653/v1/2025.findings-emnlp.792",
    pages = "14665--14688",
    ISBN = "979-8-89176-335-7"
}

@inproceedings{hallu-calib,
author = {Kalai, Adam Tauman and Vempala, Santosh S.},
title = {Calibrated Language Models Must Hallucinate},
year = {2024},
isbn = {9798400703836},
publisher = {Association for Computing Machinery},
address = {New York, NY, USA},
url = {https://doi.org/10.1145/3618260.3649777},
doi = {10.1145/3618260.3649777},
booktitle = {Proceedings of the 56th Annual ACM Symposium on Theory of Computing},
pages = {160–171},
numpages = {12},
location = {Vancouver, BC, Canada},
series = {STOC 2024}
}

@inproceedings{moskvoretskii-etal-2025-adaptive,
    title = "Adaptive Retrieval Without Self-Knowledge? Bringing Uncertainty Back Home",
    author = "Moskvoretskii, Viktor  and
      Marina, Maria  and
      Salnikov, Mikhail  and
      Ivanov, Nikolay  and
      Pletenev, Sergey  and
      Galimzianova, Daria  and
      Krayko, Nikita  and
      Konovalov, Vasily  and
      Nikishina, Irina  and
      Panchenko, Alexander",
    editor = "Che, Wanxiang  and
      Nabende, Joyce  and
      Shutova, Ekaterina  and
      Pilehvar, Mohammad Taher",
    booktitle = "Proceedings of the 63rd Annual Meeting of the Association for Computational Linguistics (Volume 1: Long Papers)",
    month = jul,
    year = "2025",
    address = "Vienna, Austria",
    publisher = "Association for Computational Linguistics",
    url = "https://aclanthology.org/2025.acl-long.319/",
    doi = "10.18653/v1/2025.acl-long.319",
    pages = "6355--6384",
    ISBN = "979-8-89176-251-0"
}

@inproceedings{wiemerslage-wense-2025-improving,
    title = "Improving Low-Resource Morphological Inflection via Self-Supervised Objectives",
    author = "Wiemerslage, Adam  and
      Wense, Katharina Von Der",
    editor = "Che, Wanxiang  and
      Nabende, Joyce  and
      Shutova, Ekaterina  and
      Pilehvar, Mohammad Taher",
    booktitle = "Proceedings of the 63rd Annual Meeting of the Association for Computational Linguistics (Volume 1: Long Papers)",
    month = jul,
    year = "2025",
    address = "Vienna, Austria",
    publisher = "Association for Computational Linguistics",
    url = "https://aclanthology.org/2025.acl-long.1195/",
    doi = "10.18653/v1/2025.acl-long.1195",
    pages = "24494--24510",
    ISBN = "979-8-89176-251-0"
}

@inproceedings{durrani-etal-2025-editing,
    title = "Editing Across Languages: A Survey of Multilingual Knowledge Editing",
    author = "Durrani, Nadir  and
      Mousi, Basel  and
      Dalvi, Fahim",
    editor = "Christodoulopoulos, Christos  and
      Chakraborty, Tanmoy  and
      Rose, Carolyn  and
      Peng, Violet",
    booktitle = "Proceedings of the 2025 Conference on Empirical Methods in Natural Language Processing",
    month = nov,
    year = "2025",
    address = "Suzhou, China",
    publisher = "Association for Computational Linguistics",
    url = "https://aclanthology.org/2025.emnlp-main.803/",
    doi = "10.18653/v1/2025.emnlp-main.803",
    pages = "15906--15918",
    ISBN = "979-8-89176-332-6"
}

@misc{Kadavath2022,
  title={Language Models (Mostly) Know What They Know},
  author={Saurav Kadavath and Tom Conerly and Amanda Askell and Tom Henighan and Dawn Drain and Ethan Perez and Nicholas Schiefer and Zac Hatfield-Dodds and Nova DasSarma and Eli Tran-Johnson and Scott Johnston and Sheer El-Showk and Andy Jones and Nelson Elhage and Tristan Hume and Anna Chen and Yuntao Bai and Sam Bowman and Stanislav Fort and Deep Ganguli and Danny Hernandez and Josh Jacobson and Jackson Kernion and Shauna Kravec and Liane Lovitt and Kamal Ndousse and Catherine Olsson and Sam Ringer and Dario Amodei and Tom Brown and Jack Clark and Nicholas Joseph and Ben Mann and Sam McCandlish and Chris Olah and Jared Kaplan},
  year={2022},
  eprint={2207.05221},
  archivePrefix={arXiv},
  primaryClass={cs.CL},
  url={https://arxiv.org/abs/2207.05221},
}

@misc{Wang2026SKR,
  title={Self-Knowledge Re-expression: A Fully Local Method for Adapting LLMs to Tasks Using Intrinsic Knowledge},
  author={Mengyu Wang and Xiaoying Zhi and Zhiyi Li and Robin Schmucker and Shay B. Cohen and Tiejun Ma and Fran Silavong},
  year={2026},
  eprint={2604.22939},
  archivePrefix={arXiv},
  primaryClass={cs.CL},
  url={https://arxiv.org/abs/2604.22939},
}

\appendix

\section{Model and Inference Details}
\label{app:models}
Models were accessed through the official OpenAI and Anthropic Python SDKs using their public APIs. All experiments used deterministic decoding with temperature set to 0. Maximum output length was capped at 450 tokens. Remaining provider-specific generation parameters were left at default values. Table~\ref{tab:model_details} summarises the inference settings used throughout all experiments.

\begin{table}[h]
\centering
\small

\begin{tabular}{lll}
\toprule
Setting & GPT-5.4 & Claude Sonnet 4.6 \\
\midrule
Provider & OpenAI & Anthropic \\
Access Method & Python SDK/API & Python SDK/API \\
Temperature & 0.0 & 0.0 \\
Decoding & Deterministic & Deterministic \\
Other Parameters & Default & Default \\
\bottomrule
\end{tabular}
\caption{Model access and inference settings.}
\label{tab:model_details}
\end{table}

\section{Marathi Dataset Translation}
\label{app:marathi-glo}
TriviaQA, RiddleBench and RealMath datasets with questions and reference answers were translated from English to Marathi using the Google Translate API. Two independent native Marathi speakers reviewed translated samples and verified semantic equivalence through spot-checking. Prompt templates were independently translated and reviewed by the same annotators. Proper nouns were transliterated into Marathi where appropriate.

\section{Marathi Local Benchmark}
\label{app:marathi-local}
\begin{figure}[h]
    \centering
    \includegraphics[width=\linewidth]{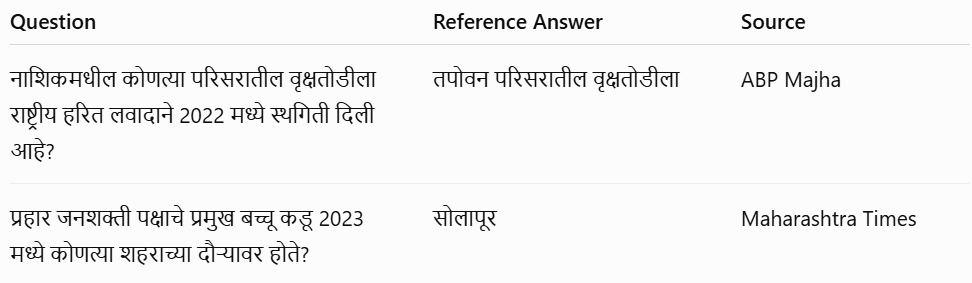}
    \caption{Examples from the Marathi-Local benchmark}
    \label{tab:marathi_examples}
\end{figure}

Figure~\ref{tab:marathi_examples} presents representative examples from the Marathi-Local benchmark. Questions were manually authored by native Marathi speakers based on region-specific events reported in Marathi-language news sources prior to model knowledge cut-off dates.

\end{document}